\documentclass[letterpaper, 10 pt, conference]{ieeeconf} 
\usepackage{amsmath} 
\usepackage{amsfonts,amssymb} 
\usepackage{comment} 
\usepackage{graphicx} 
\usepackage[labelformat=parens]{subcaption} 
\usepackage{float} 
\usepackage[inkscapelatex=false]{svg} 
\usepackage{array} 
\usepackage{makecell} 
\usepackage{caption} 
\usepackage{emf} 
\usepackage{threeparttable} 
\usepackage{booktabs} 
\usepackage{multirow} 
\usepackage{amssymb} 
\usepackage{verbatim} 
\usepackage{stfloats} 
\usepackage{cite} 
\makeatletter             
\let\NAT@parse\undefined  
\makeatother              
\usepackage{hyperref} 
\usepackage[acronym]{glossaries} 
\usepackage{afterpage}

\IEEEoverridecommandlockouts                              
\title{\LARGE \bf
Vision-Augmented On-Track System Identification for Autonomous Racing via Attention-Based Priors and Iterative Neural Correction
}

\author{
Zhiping Wu\textsuperscript{*}, 
Cheng Hu\textsuperscript{*}, 
Yiqin Wang,
Lei Xie\textsuperscript{†}, 
and Hongye Su
\thanks{\textsuperscript{*}These authors contributed equally to this work.}
\thanks{\textsuperscript{†}The corresponding author of this paper.}
\thanks{Zhiping Wu, Cheng Hu, Yiqin Wang, Lei Xie, and Hongye Su are with the State Key Laboratory of Industrial, Zhejiang University, Hangzhou 310027, China. Emails: \{wuzhiping, 22032081, yiqinwang\}@zju.edu.cn; \{leix, hysu\}@iipc.zju.edu.cn.}
}

\let\oldtwocolumn\twocolumn
\renewcommand\twocolumn[1][]{%
    \oldtwocolumn[{#1}{
    \begin{center}
           \includegraphics[width=0.95\textwidth]{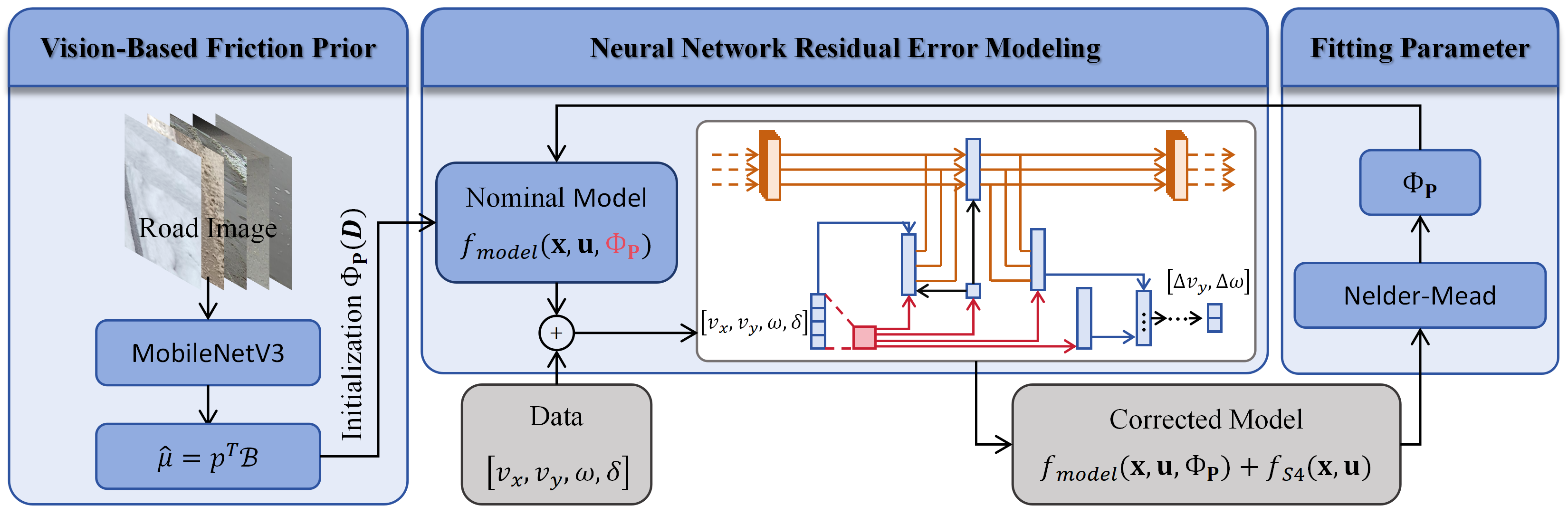}
           \captionof{figure}{The proposed vision-augmented, iterative closed-loop system identification framework. A MobileNetV3-based vision module provides a heuristic friction prior to warm-start the physical model's peak friction parameter ($D$). Concurrently, the \gls{s4} network captures high-frequency dynamic residuals by learning long-range dependencies from the state and input data. Finally, the derivative-free Nelder-Mead algorithm extracts physically bounded tire parameters ($\Phi_\mathbf{p}$) via hybrid virtual simulation, enabling continuous iterative refinement of the nominal model.}
           \label{fig:overall_architecture}
        \end{center}
    }]
}

\begin{document}

\newacronym{mpc}{MPC}{Model Predictive Control}
\newacronym{trfc}{TRFC}{Tire-Road Friction Coefficient}
\newacronym{cog}{CoG}{Center of Gravity}
\newacronym{pp}{PP}{Pure Pursuit}

\newacronym{mlp}{MLP}{Multi-Layer Perceptron}
\newacronym{mlps}{MLPs}{Multi-Layer Perceptrons}
\newacronym{rnn}{RNN}{Recurrent Neural Network}
\newacronym{rnns}{RNNs}{Recurrent Neural Networks}
\newacronym{s4}{S4}{Structured State Space Sequence}
\newacronym{cnn}{CNN}{Convolutional Neural Network}
\newacronym{cnns}{CNNs}{Convolutional Neural Networks}
\newacronym{se}{SE}{Squeeze-and-Excitation}
\newacronym{hippo}{HiPPO}{High-order Polynomial Projection Operator}
\newacronym{gelu}{GELU}{Gaussian Error Linear Unit}

\newacronym{nls}{NLS}{Non-Linear Least Squares}
\newacronym{ekf}{EKF}{Extended Kalman Filter}
\newacronym{ckf}{CKF}{Cubature Kalman Filter}
\newacronym{nm}{NM}{Nelder-Mead}
\newacronym{sgd}{SGD}{Stochastic Gradient Descent}

\newacronym{lti}{LTI}{Linear Time-Invariant}
\newacronym{zoh}{ZOH}{Zero-Order Hold}
\newacronym{fft}{FFT}{Fast Fourier Transform}

\newacronym{rscd}{RSCD}{Road Surface Condition Dataset}
\newacronym{rmse}{RMSE}{Root Mean Square Error}
\newacronym{flops}{FLOPs}{Floating-Point Operations}

\maketitle

\thispagestyle{empty}
\pagestyle{empty}

\begin{abstract}
Operating autonomous vehicles at the absolute limits of handling requires precise, real-time identification of highly non-linear tire dynamics. However, traditional online optimization methods suffer from “cold-start" initialization failures and struggle to model high-frequency transient dynamics. To address these bottlenecks, this paper proposes a novel vision-augmented, iterative system identification framework. First, a lightweight \gls{cnn} (MobileNetV3) translates visual road textures into a continuous heuristic friction prior, providing a robust "warm-start" for parameter optimization. Next, a \gls{s4} model captures complex temporal dynamic residuals, circumventing the memory and latency limitations of traditional MLPs and RNNs. Finally, a derivative-free Nelder-Mead algorithm iteratively extracts physically interpretable Pacejka tire parameters via a hybrid virtual simulation. Co-simulation in CarSim demonstrates that the lightweight vision backbone reduces friction estimation error by 76.1\% using 85\% fewer FLOPs, accelerating cold-start convergence by 71.4\%. Furthermore, the \gls{s4}-augmented framework improves parameter extraction accuracy and decreases lateral force RMSE by over 60\% by effectively capturing complex vehicle dynamics, demonstrating superior performance compared to conventional neural architectures.
\end{abstract}

\section{Introduction}

Operating autonomous vehicles at the limits of handling such as in autonomous racing or emergency collision avoidance requires high-fidelity models of vehicle and tire dynamics. Since the highly non-linear tire-road interface fundamentally defines the physical boundaries of vehicle maneuverability, the precise identification of tire parameters and road friction coefficients becomes a critical prerequisite for effective motion planning and robust Model Predictive Control (MPC)\cite{fu2025residual,rosolia2017learning,li2024reduce}.

Traditionally, system identification for complex tire models relies on offline, steady-state maneuvers\cite{pacejka2005tire}. While these methods yield high-fidelity baseline models, they lack the agility to adapt to dynamic environmental stochasticity, such as sudden friction transitions. To bridge this gap, online identification frameworks leverage real-time vehicle telemetry. However, classical optimization-based solvers often struggle in high-transient racing  cenarios. Due to the highly non-convex landscape of empirical tire models, these techniques exhibit extreme sensitivity to parameter initialization. A suboptimal initial estimate frequently results in ill-conditioned Jacobian matrices and numerical instability, leading to excessive iterations that compromise real-time computational tractability.

Furthermore, even when optimization solvers converge on nominal physical parameters, purely physics-based models often fail to capture high-frequency unmodeled transients. While recent research\cite{kim2024data} has pivoted toward data-driven residual learning to bridge these structural mismatches, standard architectures exhibit inherent limitations. Multi-Layer Perceptrons (MLP) lack the capacity to account for temporal dependencies and underlying physics\cite{mohammadi2024deep}, whereas \gls{rnn} frequently encounter vanishing or exploding gradient issues\cite{tsiligkaridis2020personalized}, hindering their ability to model long-term dynamic sequences in high-speed racing.

This paper proposes a vision-augmented, iterative framework for system identification (Fig. \ref{fig:overall_architecture}). The approach utilizes a 'warm-start' mechanism to constrain the parameter search space, followed by a \gls{s4} model to capture high-frequency residuals. The process concludes by extracting Pacejka parameters via the Nelder-Mead simplex algorithm within a hybrid virtual simulation, ensuring physical consistency.

The main contributions of this paper are summarized as follows:
\begin{itemize}
    \item \textbf{Vision-Accelerated Parameter Initialization:} A methodology is introduced to map categorical visual road classifications into continuous heuristic friction priors via probabilistic mapping. This warm-start mechanism effectively eliminates transient convergence delays.
    \item \textbf{High-Frequency Residual Modeling via \gls{s4}:} This work employs the \gls{s4} architecture for vehicle dynamic residual learning, leveraging global convolutions to model long-term inertial dependencies while circumventing the sequence degradation inherent in recurrent models.
    \item \textbf{Iterative Derivative-Free Parameter Extraction:} An iterative closed-loop framework is proposed, utilizing the Nelder-Mead algorithm on hybrid virtual simulation data. This approach ensures that the identified tire models remain physically interpretable and strictly bounded.
\end{itemize}

\section{Related Work}

\subsection{Predictive vs. Reactive Friction Estimation}
Accurate estimation of the tire-road friction coefficient (TRFC) is a critical prerequisite for limit-handling control, as it defines the tractable regions of the tire friction circle. In autonomous racing, TRFC overestimation triggers excessive slip or instability, while underestimation yields overly conservative trajectories that compromise lap time\cite{pacejka2005tire}. Traditional model-based observers, such as the Extended or Cubature Kalman Filters, are fundamentally reactive, identifying friction variations only post-facto—after the vehicle's dynamic state has deviated \cite{wang2020novel}. This inherent response lag severely jeopardizes tracking stability during abrupt surface transitions at high velocities.

To mitigate this, cameras offer a crucial "look-ahead" horizon. Extracting predictive features from this visual stream traditionally relies on standard deep \gls{cnn} like ResNet \cite{he2016deep} and EfficientNet \cite{tan2019efficientnet}. However, their substantial computational footprint induces inference latency that fundamentally undermines the predictive advantage required for autonomous racing. Beyond the need for highly efficient backbones, standard image classification yields discrete labels incompatible with continuous controllers. Without physical constraints, these data-driven models are vulnerable to the sim-to-real gap and "physical hallucinations" under unfamiliar conditions\cite{ding2025understanding}. Consequently, our approach treats lightweight visual classification as a statistical preview rather than an absolute sensor. By mapping categorical probabilities to friction expectations via physical basis vectors, we extract a robust heuristic prior to "warm-start" the identification loop, effectively decoupling visual uncertainty from the final vehicle dynamic boundaries.

\subsection{On-Track System Identification at the Limits}
Accurately modeling highly non-linear tire dynamics is fundamental to autonomous racing. Traditional offline identification relies on structured experiments and optimization techniques like Non-Linear Least Squares (NLS)\cite{brunner2017repetitive}, which inherently fail to adapt to mid-race environmental changes. While online \textit{on-track} identification \cite{dikici2025learning} leverages continuous telemetry to dynamically update parameters, an inaccurate initial guess renders the optimization landscape ill-conditioned, leading to severe convergence delays. Our approach circumvents this algorithmic brittleness by coupling the proposed visual warm-start with the derivative-free Nelder-Mead simplex algorithm, ensuring robust and rapid parameter convergence.

\subsection{Hybrid Dynamics and Sequence-Based Residual Modeling}
The integration of machine learning with first-principles physics has gained significant traction for capturing complex vehicle dynamics at the edge of handling. A common paradigm utilizes a nominal physical model to capture baseline behavior, while a neural network learns the high-frequency unmodeled residual dynamics\cite{ajay2019combining,barbulescu2025physics}. Previous frameworks have utilized MLPs and RNNs for this task. However, MLPs process telemetry independently, entirely disregarding the temporal "memory" of the vehicle's inertial physics. Conversely, RNNs suffer from vanishing gradients and high computational latency due to sequential inefficiencies. Recently, \gls{s4} \cite{gu2021efficiently} models have emerged as an efficient alternative, handling long-range dependencies mathematically and empirically. Rooted in continuous-time linear systems, \gls{s4} explicitly models continuous dynamics while scaling efficiently via global convolutions.

\section{Methodology}

\subsection{Nominal Vehicle and Tire Dynamics}
\begin{figure}[htbp]
\centering 
\includegraphics[width=0.8\columnwidth]{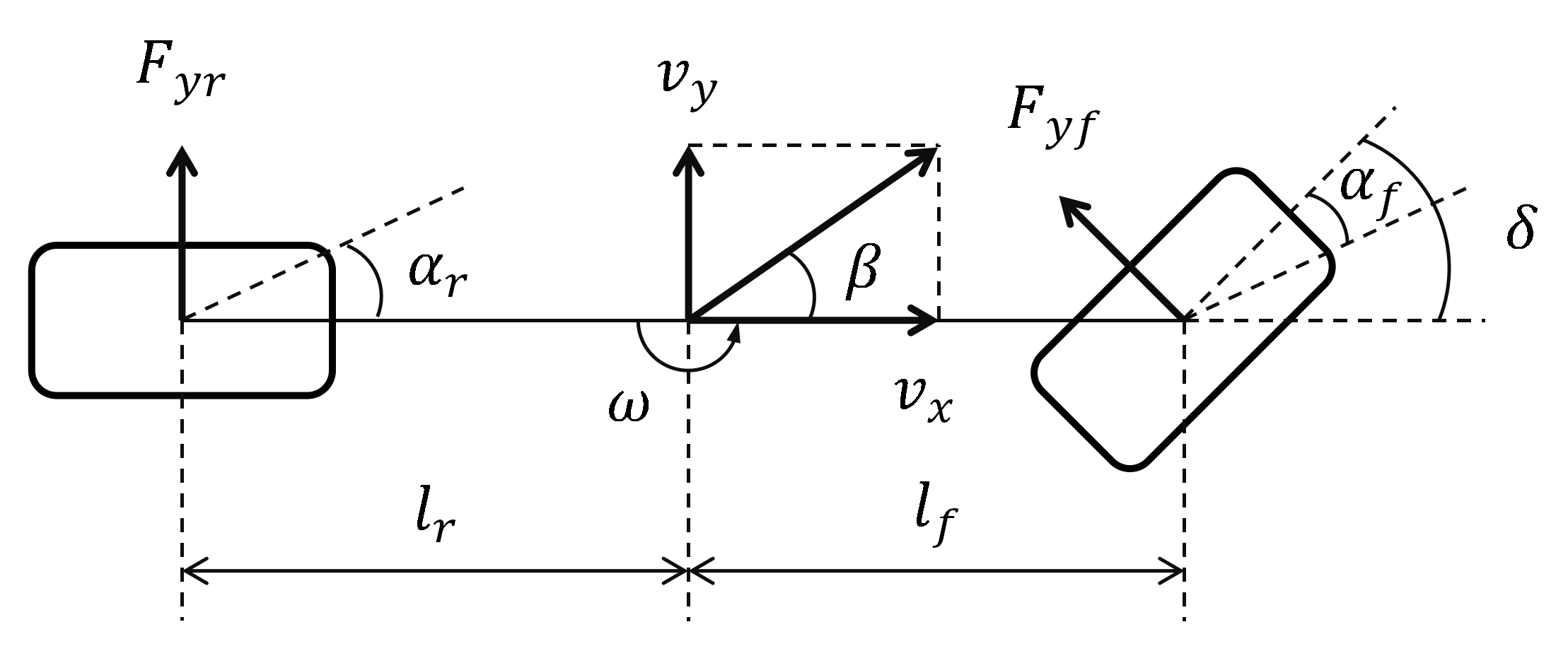}
\caption{Schematic representation of the dynamic single-track model.}
\label{fig:single_track}
\end{figure}

The vehicle dynamics are represented by the dynamic single-track model (Fig.~\ref{fig:single_track}). This model is widely utilized in autonomous racing research as it effectively captures critical tire-road interactions during both steady-state and transient cornering, while maintaining computational efficiency by decoupling longitudinal and lateral dynamics. Under the assumption of a sufficiently low \gls{cog}, the influences of longitudinal and lateral load transfers are treated as negligible.

The lateral and yaw dynamics are governed by the following equations of motion:

\begin{equation}
    \dot{v}_y = \frac{1}{m}(F_{yr} + F_{yf} \cos \delta - m v_x \omega)\label{eq:vy}
\end{equation}

\begin{equation}
    \dot{\omega} = \frac{1}{I_z}(F_{yf} l_f \cos \delta - F_{yr} l_r)
    \label{eq:omega}
\end{equation}

where $m$ is the vehicle mass and $I_z$ is the yaw moment of inertia. The state variables $v_x$, $v_y$, and $\omega$ denote longitudinal velocity, lateral velocity, and yaw rate, respectively. Geometric parameters $l_f$ and $l_r$ represent the distances from the \gls{cog} to the front and rear axles. The control input is the steering angle $\delta$, while $F_{yf}$ and $F_{yr}$ are the lateral tire forces.

These lateral forces are functions of the tire slip angles, $\alpha_f$ and $\alpha_r$, kinematically derived as:

\begin{equation}
    \alpha_f = \delta - \arctan\left(\frac{v_y + l_f \omega}{v_x}\right)\label{eq:alpha_f}
\end{equation}

\begin{equation}
    \alpha_r = -\arctan\left(\frac{v_y - l_r \omega}{v_x}\right)\label{eq:alpha_r}
\end{equation}

To model highly non-linear interactions at the handling limits, the Pacejka Magic Formula is employed:

\begin{equation}
    F_{yi} = D_i \sin\Big(C_i \arctan\big(B_i\alpha_i - E_i(B_i\alpha_i - \arctan(B_i\alpha_i))\big)\Big)\label{eq:pacejka}
\end{equation}

for $i \in \{f, r\}$. The empirical tire behavior is defined by the parameter vector $\Phi_{p,i} = [B_i, C_i, D_i, E_i]$, representing stiffness, shape, peak friction, and curvature, respectively.

A major challenge in real-time identification of Eq. \eqref{eq:pacejka} is its sensitivity to initialization. To mitigate the risk of local minima, our framework introduces a coupling mechanism: the peak friction parameter $D$ is initialized using a vision-based heuristic prior ($\mu_{prior}$), the derivation of which is detailed in the following section.

\subsection{Vision-Based Friction Initialization}

\subsubsection{Dataset Foundation and Preprocessing}
To establish a reliable visual foundation for friction estimation, we leverage the Road Surface Condition Dataset (RSCD) \cite{zhao2023comprehensive}. Spanning approximately 700 km across diverse environments, the RSCD encompasses over 1 million image samples with high-fidelity labels for friction levels, surface unevenness, and material properties. This extensive scale is instrumental in ensuring robust generalization across the intricate visual textures encountered in autonomous racing.

To bridge the gap between categorical deep learning outputs and continuous physical control variables, we adopt a probabilistic mapping framework to translate high-dimensional visual features into numerical friction coefficients \cite{otoofi2023estimating}. Given the distinct physical characteristics of racing tracks, we define a set of $N$ macro-classes, formally associated with a predefined physical friction basis vector $\mathcal{B} \in \mathbb{R}^N$:

\begin{equation}
    \mathcal{B} = [\mu_1, \mu_2, \dots, \mu_N]^T \label{eq:basis_vector}
\end{equation}

where each element $\mu_i$ corresponds to the nominal friction coefficient associated with the $i$-th surface category.

During the forward pass, the \gls{cnn} backbone extracts features to generate a raw logit vector $\mathbf{z} \in \mathbb{R}^N$. This vector is transformed into a categorical probability distribution $\mathbf{p}$ via the Softmax function:

\begin{equation}
    \mathbf{p}_i = \frac{\exp(\mathbf{z}_i)}{\sum_{j=1}^{N} \exp(\mathbf{z}_j)} 
    \label{eq:softmax_prob}
\end{equation}

The estimated friction coefficient $\hat{\mu}$ is then derived as the weighted expectation over the physical bases:

\begin{equation}
    \hat{\mu} = \sum_{i=1}^{N} \mathbf{p}_i \cdot \mu_i = \mathbf{p}^T \mathcal{B}
    \label{eq:expected_friction}
\end{equation}

This formulation yields a differentiable mapping that smooths transitions between discrete surface states. By converting categorical confidence into a continuous prior, the framework provides numerically stable inputs for subsequent online system identification, enabling the vehicle to adapt its internal dynamic models in real-time.

\subsubsection{\gls{cnn} Backbone}
\begin{figure*}[htbp]
\centerline{\includegraphics[width=\textwidth]{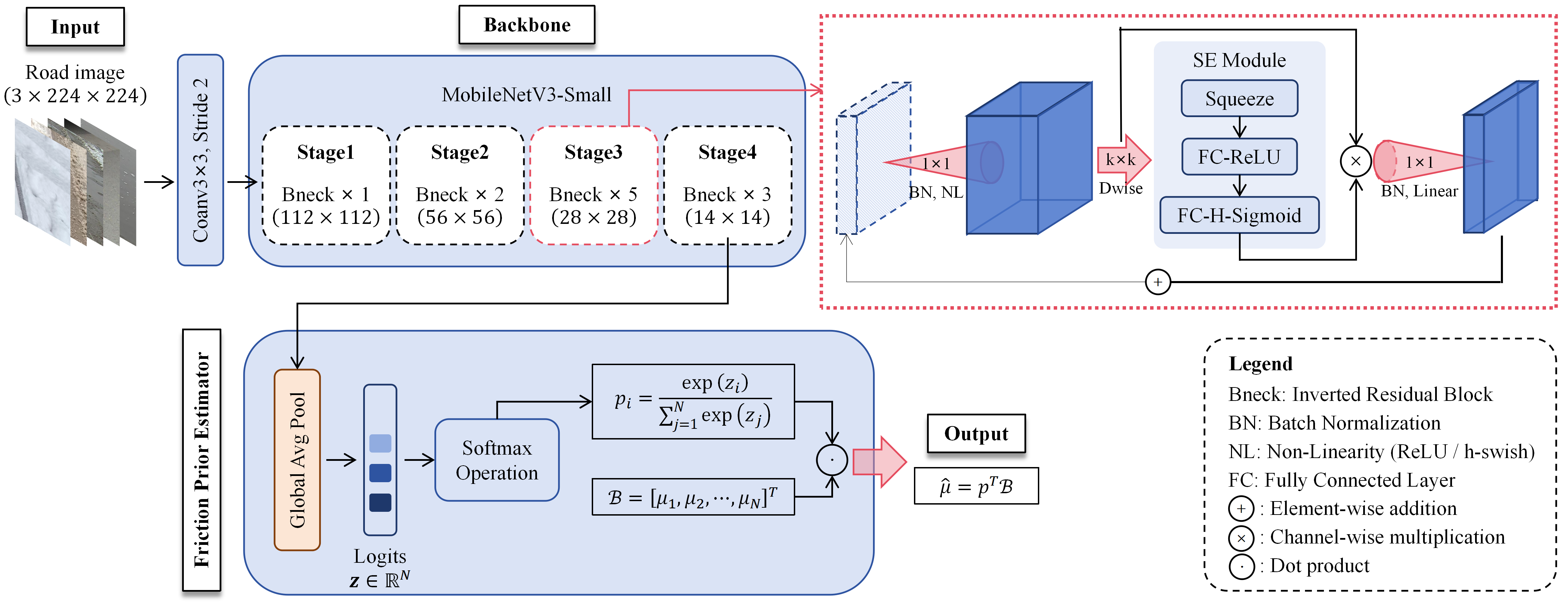}} 
\caption{Architecture of the vision-based friction estimation module. The system utilizes a MobileNetV3-Small backbone across four hierarchical stages for efficient feature extraction. The Inverted Residual Block topology incorporates a \gls{se} attention mechanism for texture recalibration. The Probabilistic Friction Mapper executes a classification-to-regression mapping, weighting the Softmax distribution $\mathbf{p}$ by the physical basis vector $\mathcal{B}$ to derive the continuous expected friction coefficient $\hat{\mu}$.}
\label{fig:Mobilenet}
\end{figure*}

To meet the rigorous real-time constraints of high-speed autonomous racing, we employ MobileNetV3-Small \cite{howard2019searching} as the vision backbone (Fig.~\ref{fig:Mobilenet}). This selection is motivated by the requirement for deterministic, low-latency feature extraction, ensuring that visual priors remain synchronized with the vehicle's high-frequency control loop and the subsequent system identification stage.

The efficiency of MobileNetV3-Small stems from its Inverted Residual Block structure. Unlike conventional residual blocks, this architecture employs a $1 \times 1$ expansion layer to project input features into a high-dimensional latent space, followed by a depthwise convolution for spatial feature extraction. To preserve feature integrity within the high-dimensional embedding, a $1 \times 1$ linear bottleneck projects the features back to a lower-dimensional space without non-linear activation.

To enhance sensitivity to subtle road textures—such as moisture patterns or grain variations—the deeper stages integrate the \gls{se} attention mechanism. By dynamically recalibrating channel-wise responses through global context modeling, the SE module enables the network to prioritize salient surface reflections while suppressing environmental noise. This lightweight, attention-augmented configuration ensures robust prior generation within a millisecond-scale inference window, satisfying the stringent constraints of edge deployment.

\subsubsection{Heuristic Friction Mapping and the "Cold Start" Solution}
A primary challenge in real-time system identification is the "cold-start" problem, where suboptimal tire parameter initialization frequently leads to numerical divergence or entrapment in local minima. To bridge the gap between visual perception and continuous vehicle dynamics, the proposed vision module incorporates a heuristic friction mapping layer.

This layer translates the \gls{cnn} output into a nominal friction value, $\mu_{prior}$, based on standardized physical benchmarks. This vision-derived prior directly serves as the initialization seed for the peak friction parameter $D$ in the Pacejka Magic Form

\begin{equation}
    D_{initial} = \hat{\mu} = \mathbf{p}^T \mathcal{B} \label{eq:d_initial}
\end{equation}

By "warm-starting" the identification process with this visual prior, the parameter search space is significantly constrained. This strategy accelerates the convergence rate of subsequent optimization stages and enhances algorithmic robustness under extreme racing conditions.

However, while this vision-augmented initialization effectively identifies nominal physical parameters, unmodeled high-frequency transients and complex residuals inevitably persist at the absolute handling limits. To compensate for these non-linear discrepancies, a sequence-based neural correction mechanism is introduced in the following subsection.

\subsection{\gls{s4}-Based Dynamic Residual Error Modeling}

\subsubsection{Mathematical Formulation of the \gls{s4} Model}
The foundation of the \gls{s4} model is rooted in continuous-time linear time-invariant (LTI) state-space representations. It continuously maps a 1-D input signal $u(t) \in \mathbb{R}$ to an output signal $y(t) \in \mathbb{R}$ via an implicit $N$-dimensional latent state $h(t) \in \mathbb{R}^N$:

\begin{equation}
    \dot{h}(t) = \mathbf{A}h(t) + \mathbf{B}u(t), \quad y(t) = \mathbf{C}h(t) + \mathbf{D}u(t) 
    \label{eq:s4_cont}
\end{equation}

To process discrete vehicle telemetry sampled at intervals $T_s$, the continuous parameters $(\mathbf{A}, \mathbf{B})$ are discretized using a learnable step size $\Delta$. Under the Zero-Order Hold (ZOH) assumption, the continuous system is transformed into a discrete representation where the state transition is governed by $\exp(\Delta \mathbf{A})$. To maintain the physical integrity of the mapping during kernel construction, the output matrix $\mathbf{C}$ is scaled to account for the integration of the state over the discrete interval:

\begin{equation}
    \mathbf{\bar{C}} = \mathbf{C}(\exp(\Delta \mathbf{A}) - \mathbf{I})\mathbf{A}^{-1} 
    \label{eq:s4_zoh_c}
\end{equation}

Consequently, the discrete linear recurrence can be expressed as:

\begin{equation}
    h_k = \mathbf{\bar{A}}h_{k-1} + \mathbf{\bar{B}}u_k, \quad y_k = \mathbf{\bar{C}}h_k + \mathbf{D}u_k 
    \label{eq:s4_rec}
\end{equation}

where $\mathbf{\bar{A}} = \exp(\Delta \mathbf{A})$. Unlike \gls{rnn} that suffer from sequential bottlenecks and vanishing gradients, the \gls{s4} model mathematically unrolls this recurrence into a global discrete convolution. For a sequence of length $L$, the output is computed with high parallel efficiency via the Fast Fourier Transform (FFT) utilizing a structured convolutional kernel $\mathbf{\bar{K}}$:

\begin{equation}
    y = \mathbf{\bar{K}} * u, \quad \text{where} \quad \mathbf{\bar{K}}_k = \mathbf{\bar{C}} \exp(\Delta \mathbf{A})^k \mathbf{B} 
    \label{eq:s4_conv}
\end{equation}

The critical innovation of the \gls{s4} architecture lies in the specific parameterization of the state transition matrix $\mathbf{A}$. To effectively capture long-term historical dependencies—such as tire relaxation lengths and transient load transfers—$\mathbf{A}$ is structured as a diagonal matrix with complex eigenvalues and initialized utilizing the \gls{hippo} framework. The real components of these eigenvalues govern the exponential decay of past state memory, while the imaginary components characterize the oscillatory modes of the vehicle dynamics. This formulation provides a physically intuitive and mathematically rigorous mechanism for modeling the frequency-rich, non-linear transients inherent in autonomous racing.

\subsubsection{Network Architecture}
\begin{figure}[htbp]
\centerline{\includegraphics[width=\columnwidth]{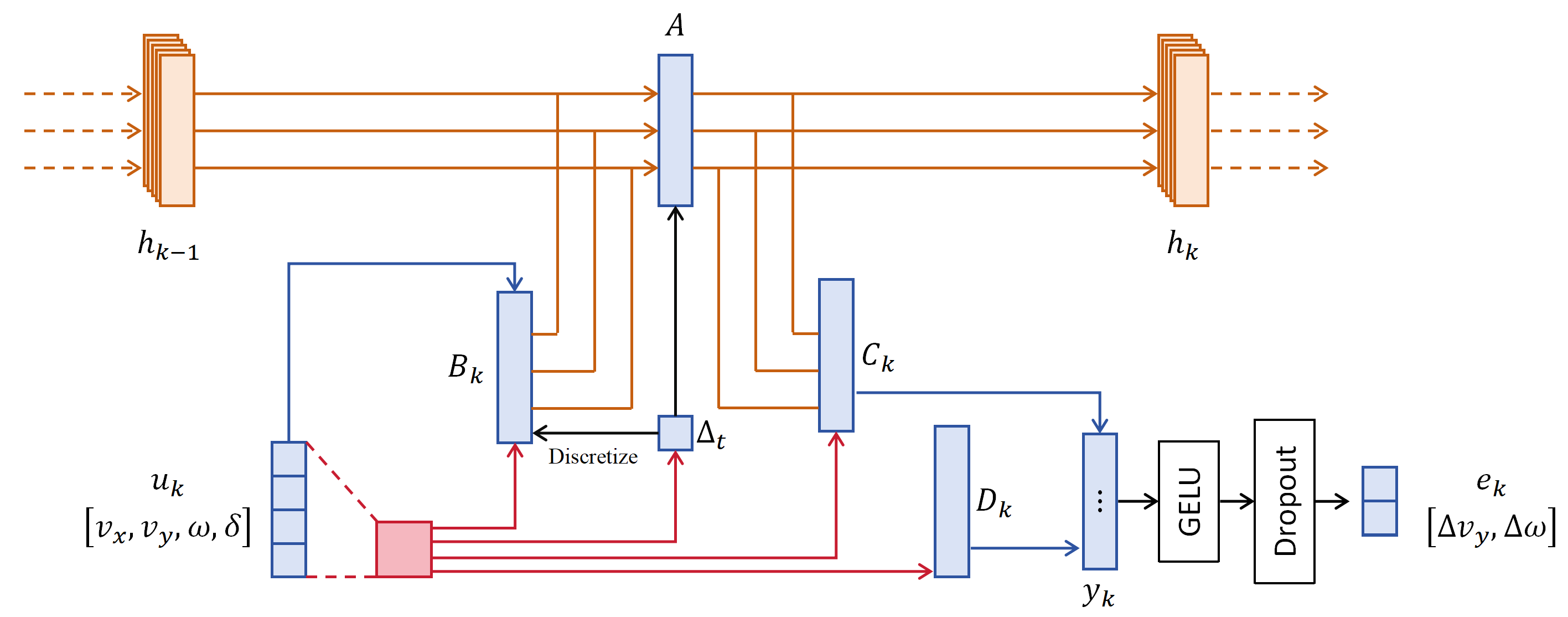}}
\caption{Architecture of the \gls{s4} model. The diagram illustrates the transition from \gls{hippo}-initialized continuous state-space to a discrete convolutional representation, enabling parallelized computation of vehicle dynamic residuals.}
\label{fig:s4_architecture}
\end{figure}

The architecture of the \gls{s4}-based residual network is illustrated in Fig. \ref{fig:s4_architecture}. The \gls{s4}-based residual network is architected to balance representational capacity with computational tractability, ensuring compliance with the real-time constraints of onboard inference. The input vector, comprising state-action pairs $u_k = [v_{x,k}, v_{y,k}, \omega_k, \delta_k]^T$, is initially projected into a high-dimensional latent space $d_{model}$ via a linear encoding layer. This latent sequence is subsequently processed by the \gls{s4} kernel through global convolutions, which extract temporal dependencies and dynamic cross-correlations across the entire time horizon $L$ without the overhead of sequential recurrence.

To ensure robust generalization on sparse on-track datasets, the \gls{s4} layer output is regularized using a \gls{gelu} activation and a dropout layer. Finally, a decoding module maps the latent representations back to the dynamic residual space $e_k = [\Delta v_y, \Delta \omega]^T$. By integrating this memory-augmented architecture, the proposed methodology overcomes the temporal amnesia typical of traditional MLPs, effectively filtering measurement noise and capturing the high-frequency transients essential for handling at the vehicle's physical limits.

\subsection{Parameter Identification with the \gls{s4}-Corrected Vehicle Model}
Dynamic track conditions inevitably induce a model mismatch ($e_k$) between the vehicle's true state $x_{k+1}$ and the nominal prediction $\hat{x}_{k+1}$ . To capture these unmodeled transients, we construct a "Corrected Vehicle Model"  that augments the nominal physics with the \gls{s4}-derived residual sequence($u_{seq})$:

\begin{equation}
    x_{k+1} = f_{model}(x_k, u_k, \Phi_p) + f_{S4}(u_{seq}) \label{eq:corrected_model}
\end{equation}

Within this virtual environment, the vehicle is simulated at a constant average longitudinal velocity, and a linear steering sweep maneuver is applied. This controlled, virtual maneuver safely forces the tires into their non-linear saturation regions. Assuming quasi-steady-state conditions ($\dot{v}_y \approx 0, \dot{\omega} \approx 0$), the true lateral tire forces can be analytically derived from the corrected states:

\begin{equation}
    F_{yr} = \frac{m l_f}{l_f + l_r} v_x \omega, \quad F_{yf} = \frac{m l_r}{l_f + l_r} \frac{v_x \omega}{\cos(\delta)} \label{eq:forces_steady}
\end{equation}

To fit the Pacejka Magic Formula to this synthesized $(F_y, \alpha)$ dataset, we employ the derivative-free Nelder-Mead algorithm. Unlike gradient-based solvers (e.g., NLS) that often diverge due to ill-conditioned Jacobian matrices in non-convex landscapes, Nelder-Mead ensures robust convergence within strict physical bounds.

%Finally, because initial model mismatches can be severe, a single extraction is insufficient. We implement a closed-loop iterative framework: the identified Pacejka parameters $\Phi_p$ update the nominal baseline, and the \gls{s4} network is retrained on the diminished residuals. This cycle (Residual Learning $\rightarrow$ Virtual Simulation $\rightarrow$ Parameter Extraction) repeats until empirical convergence. By co-evolving the physical parameters and the neural network, we effectively mitigate overfitting and yield a highly robust, interpretable tire model for real-time control.

\begin{figure}[htbp]
\centerline{\includegraphics[width=\columnwidth]{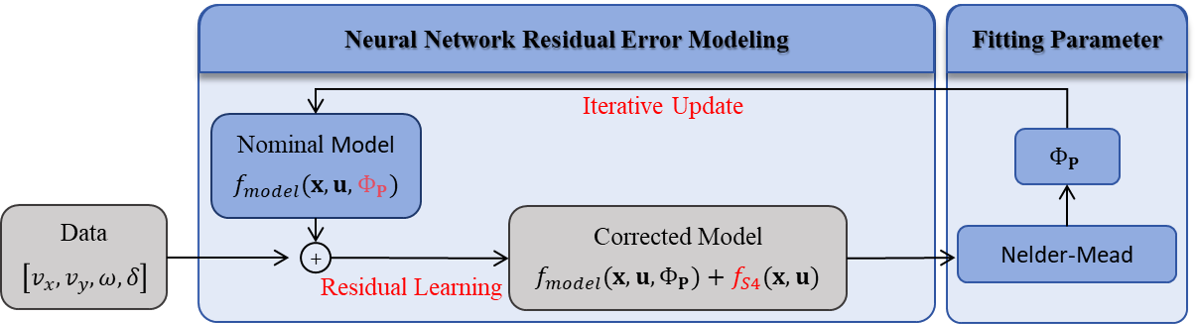}}
\caption{Iterative framework for tire model identification.}
\label{fig:iterative_franmework}
\end{figure}

Finally, since the initial model mismatch can be significant, a single extraction step is insufficient. We therefore propose a closed-loop iterative framework, where the identified Pacejka parameters $\Phi_p$ update the nominal baseline, and the \gls{s4} network is retrained on the progressively reduced residuals. This cycle (Residual Learning $\rightarrow$ Virtual Simulation $\rightarrow$ Parameter Extraction) is repeated until convergence, as illustrated in Fig.~\ref{fig:iterative_franmework}. By jointly evolving the physical parameters and the neural network, the framework mitigates overfitting and yields a robust tire model.

\section{Simulation Results}
The entire experiment is conducted on a joint co-simulation platform using MATLAB-R2023b and the high-fidelity vehicle simulation software CarSim-2019. The simulation is executed on an Intel Xeon Platinum processor and an NVIDIA RTX 4090 GPU, where the GPU is used to accelerate neural network inference.

\subsection{Simulation Setup}

To evaluate the performance of the proposed algorithm, simulation experiments are conducted in CarSim, a high-fidelity vehicle dynamics simulator designed to minimize the sim-to-real gap. A key advantage of the CarSim environment is the ability to define custom road surfaces with specific friction coefficients ($\mu$), allowing for rigorous testing across diverse track conditions.Prior to the dynamic simulations, the lightweight MobileNetV3 vision backbone is trained offline using a predefined dataset of diverse road surface images. 

The data collection and processing workflow is organized into two distinct phases: first, 30 seconds of telemetry data $[v_x, v_y, \omega, \delta]$ is collected as the vehicle follows a predefined trajectory using a Pure Pursuit controller\cite{betz2022autonomous} to train the \gls{s4} residual network ; subsequently, an online system identification process is triggered to rapidly update the Pacejka tire parameters ($\Phi_p$) through a closed-loop iterative process, ensuring the nominal model matches the refined dynamics captured by the \gls{s4}-corrected model.

\begin{figure}[htbp]
\centerline{\includegraphics[width=0.8\columnwidth,height=0.6\columnwidth]{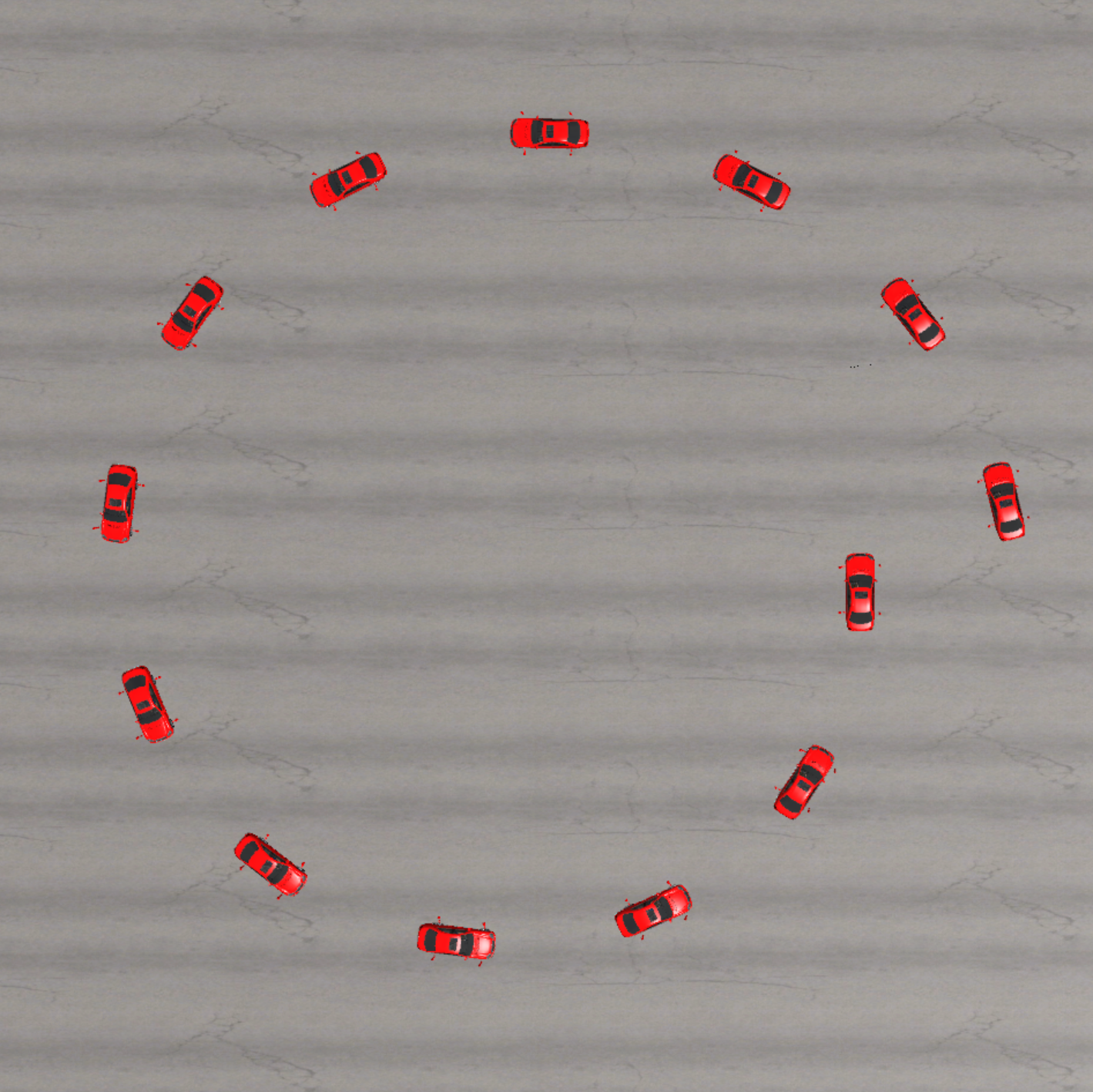}}
\caption{Visualization of the data acquisition phase in the CarSim environment.}
\label{fig:carsim_data_collection}
\end{figure}

\subsection{Results}
The vision backbone is first evaluated for friction estimation, followed by an analysis of its role in cold-start identification. The impact of network architecture on parameter extraction is then examined.

\subsubsection{Vision Backbone Evaluation}

Prior to evaluating the closed-loop vehicle dynamics, we benchmark the performance of the vision-based friction estimation module. We compare three prominent architectures—MobileNetV3-Small, ResNet-18, and EfficientNet-B0—as detailed in Table~\ref{tab:model_comparison}.

To ensure a fair and comprehensive evaluation, all three vision backbones were trained under identical experimental conditions. The input images were resized to $224 \times 224$ pixels and normalized using standard ImageNet statistics. To enhance model robustness and mitigate overfitting, we applied a series of data augmentation techniques during the training phase, including random horizontal flipping, color jittering, and random rotations between $-4^\circ$ and $4^\circ$.

The networks were trained for 40 epochs with a batch size of 256, utilizing the Stochastic Gradient Descent (SGD) optimizer. We set the momentum to 0.9, weight decay to 0.0001, and an initial learning rate of 0.05. A step learning rate scheduler was employed, decaying the learning rate by a factor of 0.1 every 15 epochs. During inference, the final friction coefficient is calculated as defined in Equation~\ref{eq:expected_friction}.

\begin{table}[htbp]
  \centering
  \caption{Quantitative comparison of candidate vision backbones.}
  \label{tab:model_comparison}
  \setlength{\tabcolsep}{3pt}
  \begin{tabular}{lccc}
    \hline
    \textbf{Metric} & \textbf{MobileNetV3 (ours)} & \textbf{ResNet-18}\cite{he2016deep} & \textbf{EfficientNet} \cite{howard2019searching}\\
    \hline
    Params (M)      & \textbf{1.52}                 & 11.18              & 4.01                  \\
    FLOPs (G)       & \textbf{0.12}                 & 3.65               & 0.83                  \\
    RMSE            & \textbf{0.102}                & 0.427              & 0.310                 \\
    Latency (ms)    & 5.91                 & \textbf{5.33}               & 8.35                  \\
    \hline
  \end{tabular}
\begin{tablenotes}
\footnotesize
\item \textit{Note:} Params (M) denotes the number of trainable parameters in millions. FLOPs (G) represents the number of floating-point operations in billions for a single forward pass. Accuracy (\%) denotes the percentage of predictions within a predefined friction coefficient error threshold. RMSE represents the root mean square error of the estimated friction coefficient, respectively. Latency (ms) denotes the average inference time per sample.
\end{tablenotes}
\end{table}

MobileNetV3 achieves the lowest RMSE of 0.102, significantly outperforming ResNet-18 and EfficientNet-B0. Specifically, compared to ResNet-18, MobileNetV3 reduces the RMSE by 0.325 (76.1\%), demonstrating substantially improved friction estimation accuracy. Compared to EfficientNet-B0, MobileNetV3 achieves a reduction of 0.208 (67.1\%). In addition to its superior accuracy, MobileNetV3 is also considerably more efficient, requiring 86.4\% fewer parameters and 96.7\% fewer FLOPs than ResNet-18, and 62.1\% fewer parameters and 85.5\% fewer FLOPs than EfficientNet-B0. Although ResNet-18 has slightly lower latency (5.33 ms vs. 5.91 ms), its significantly higher RMSE makes it unsuitable for accurate friction estimation. EfficientNet-B0, while more accurate than ResNet-18, suffers from both higher latency and substantially higher RMSE compared to MobileNetV3. Therefore, MobileNetV3 provides the best balance between accuracy and efficiency, delivering reliable millisecond-scale inference performance for autonomous racing applications.

\subsubsection{Impact of Vision Prior on Cold-Start Identification}
In practical autonomous racing, vehicles frequently enter an unseen circuit where the global TRFC is not accurately known a priori. Traditional online parameter identification algorithms must rely on a default, hardcoded initial guess ($\mu_0$). If this initialization heavily deviates from the actual surface conditions, the optimization solver requires extensive closed-loop iterations to converge, leading to initial erratic control behavior and compromised trajectory tracking.

To quantify the benefits of integrating a visual prior, we established an initial maneuvering experiment on a uniform yet unknown surface. To construct this experimental setup, high-resolution road texture images—corresponding to specific, known friction coefficients—were first mapped onto the Carsim simulated road. Subsequently, prior to data acquisition, the physical tire-road friction coefficient ($\mu$) defined within the CarSim environment was strictly aligned with the inherent friction value represented by the visual texture, thereby guaranteeing visual-physical consistency. We triggered the online system identification during the vehicle's initial acceleration phase to benchmark two distinct pipelines: a standard optimization process operating without visual context, and our proposed vision-augmented framework. In the latter pipeline, the rendered road texture images are directly processed by the lightweight MobileNetV3 backbone to instantly yield a heuristic friction expectation $\hat{\mu}$.

\begin{table}[t]
\centering
\caption{Effect of Vision Prior on Identification Initialization}
\label{tab:vision_prior}
{
\begin{tabular}{lcc}
\toprule
\textbf{Metric} 
& \textbf{No Vision} 
& \textbf{Vision} \\
\midrule

Init. Friction $\mu_0$ 
& 0.51 
& 0.51 \\

Vision Friction $\hat{\mu}$ 
& -- 
& 0.80 \\

$F_{yf}/F_{zf}$ RMSE 
& 0.219 
& \textbf{0.076} \\

$F_{yr}/F_{zr}$ RMSE 
& 0.073 
& \textbf{0.046} \\

Iter. 
& 7 
& \textbf{2} \\

\bottomrule
\end{tabular}
}
\end{table}

The quantitative advantages of this framework are detailed in Table~\ref{tab:vision_prior}. Both pipelines start with a severely mismatched default initial friction guess of $\mu_0 = 0.51$. In the vision-augmented pipeline, the network immediately extracts a heuristic prior of $\hat{\mu} = 0.80$ from the simulated visual textures. This "warm-start" fundamentally transforms the optimization landscape. Most notably, it accelerates solver convergence by reducing the required Nelder-Mead iterations from 7 down to just 2—a 71.4\% reduction in initial setup time. Furthermore, the vision prior drastically improves the parameter extraction accuracy. Compared to the baseline, the vision-augmented approach reduces the $F_{yf}/F_{zf}$ RMSE by 0.143 (a 65.3\% improvement) and the $F_{yr}/F_{zr}$ RMSE by 0.027 (a 37.0\% improvement).

\begin{figure}[t]
  \centering
  \begin{subfigure}{0.38\textwidth}
    \centering
    \includegraphics[width=\linewidth]{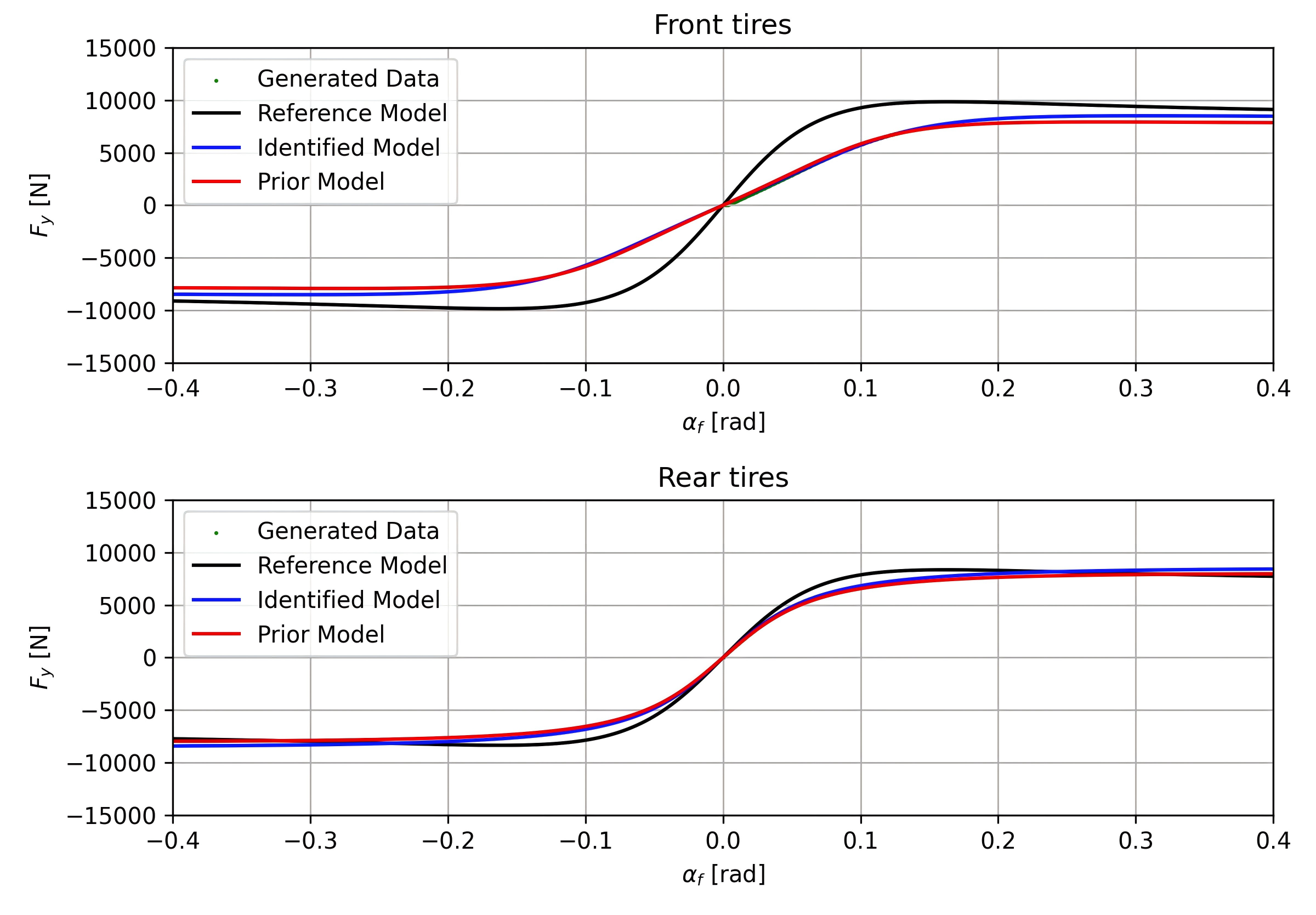}
    \caption{Without the visual prior}
    \label{fig:no_prior}
  \end{subfigure}
  \hfill
  \begin{subfigure}{0.38\textwidth}
    \centering
    \includegraphics[width=\linewidth]{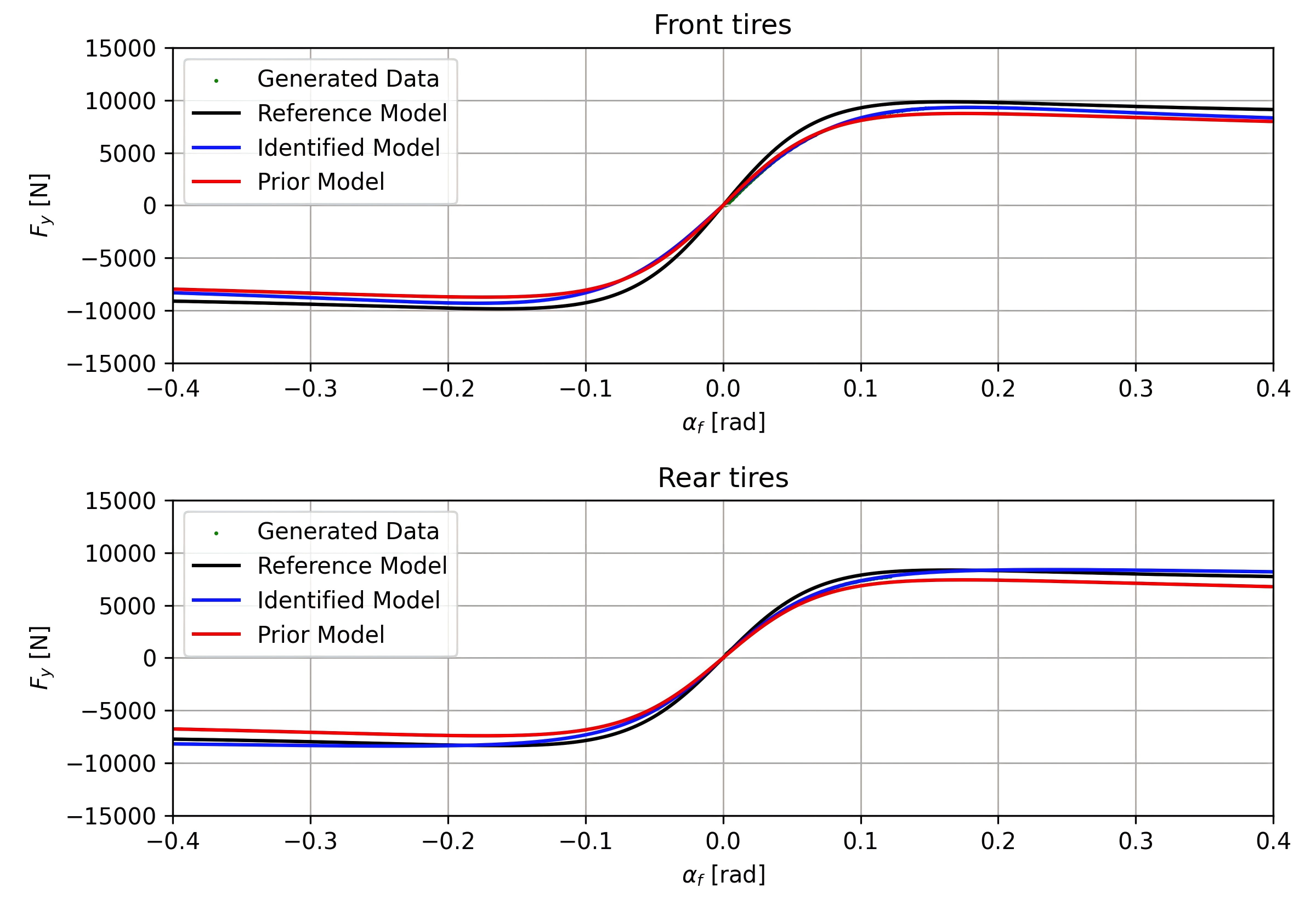}
    \caption{With the visual prior}
    \label{fig:vision_prior}
  \end{subfigure}
  \caption{Comparison of identified tire force curves with and without the vision prior. The vision prior significantly improves the initial parameter accuracy.}
  \label{fig:vision_comparison}
\end{figure}

The physical fidelity of this initialization is further corroborated by the identified tire force curves in Fig.~\ref{fig:vision_comparison}. Without the visual prior (Fig.~\ref{fig:no_prior}), the substantial initial parameter mismatch causes the solver to struggle through a prolonged search space, resulting in noticeable deviations in the estimated non-linear handling regions. Conversely, by conditioning the solver with the vision-derived expectation (Fig.~\ref{fig:vision_prior}), the framework effectively places the initial simplex directly in the vicinity of the global minimum. This capability ensures that the vehicle acquires a highly accurate and stable dynamic model almost instantly upon race initiation, effectively preventing cold-start instability.

\subsubsection{Impact of Network Architecture on Parameter Extraction}

To isolate the architectural impact on physical parameter extraction, we benchmark the proposed \gls{s4} framework against a memoryless \gls{mlp} and a \gls{rnn}. The fidelity of the extracted Pacejka models fundamentally depends on how accurately the network captures unmodeled transient dynamics during the virtual steady-state simulation. 

\begin{table}[htbp]
\caption{Performance Metrics of Different Residual Architectures for Parameter Extraction}
\label{tab:architecture_metrics}
\centering
\begin{tabular}{@{}l|cc|c@{}}
\toprule
\textbf{Architecture} & \textbf{$F_{yf}/F_{zf}$ RMSE} & \textbf{$F_{yr}/F_{zr}$ RMSE} & \textbf{Time [s]} \\ \midrule
MLP                   & 0.234                      & 0.071                      & \textbf{5.3}               \\
RNN                   & 0.097                      & 0.084             & 12.9              \\
\textbf{\gls{s4} (Ours)}    & \textbf{0.051}                      & \textbf{0.033}                      & 12.2     \\ \bottomrule
\end{tabular}
\end{table}

As summarized in Table \ref{tab:architecture_metrics}, the \gls{s4} framework achieves the lowest RMSE for both front and rear normalized lateral forces, significantly outperforming the MLP and RNN architectures. Specifically, compared to the MLP, \gls{s4} reduces the $F_{yf}/F_{zf}$ RMSE by 0.183 (a 78.2\% improvement) and the $F_{yr}/F_{zr}$ RMSE by 0.038 (53.5\%), demonstrating a substantially improved ability to capture unmodeled transient dynamics. When compared to the RNN, \gls{s4} achieves a reduction of 0.046 (47.4\%) in $F_{yf}/F_{zf}$ RMSE and 0.051 (60.7\%) in $F_{yr}/F_{zr}$ RMSE. Although the memoryless MLP offers the lowest computational time (5.3~s vs.\ 12.2~s), its significantly higher RMSE indicates a fundamental failure to account for the vehicle's temporal inertia. The RNN, while more accurate than the MLP for the front tire, suffers from sequential execution bottlenecks resulting in the highest computational time (12.9~s) alongside worse rear tire estimation. Therefore, the \gls{s4} framework provides the optimal balance between precision and efficiency, yielding highly reliable sequence-level accuracy necessary for high-fidelity Pacejka parameter identification.

\begin{figure*}[t]
  \centering
  \begin{subfigure}{0.32\textwidth}
    \centering
    \includegraphics[width=\linewidth]{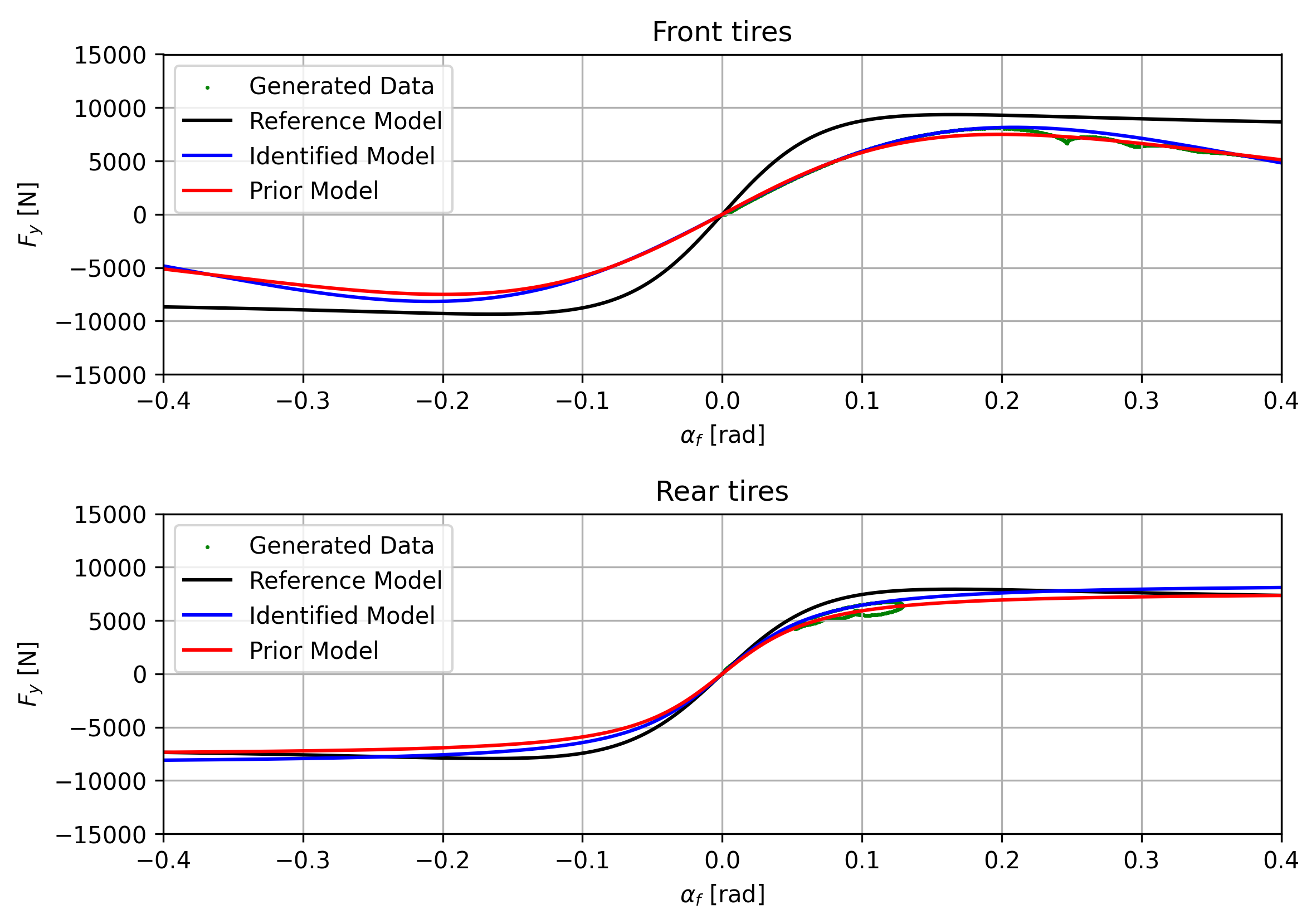}
    \caption{MLP}
    \label{fig:mlp_fit}
  \end{subfigure}
  \hfill
  \begin{subfigure}{0.32\textwidth}
    \centering
    \includegraphics[width=\linewidth]{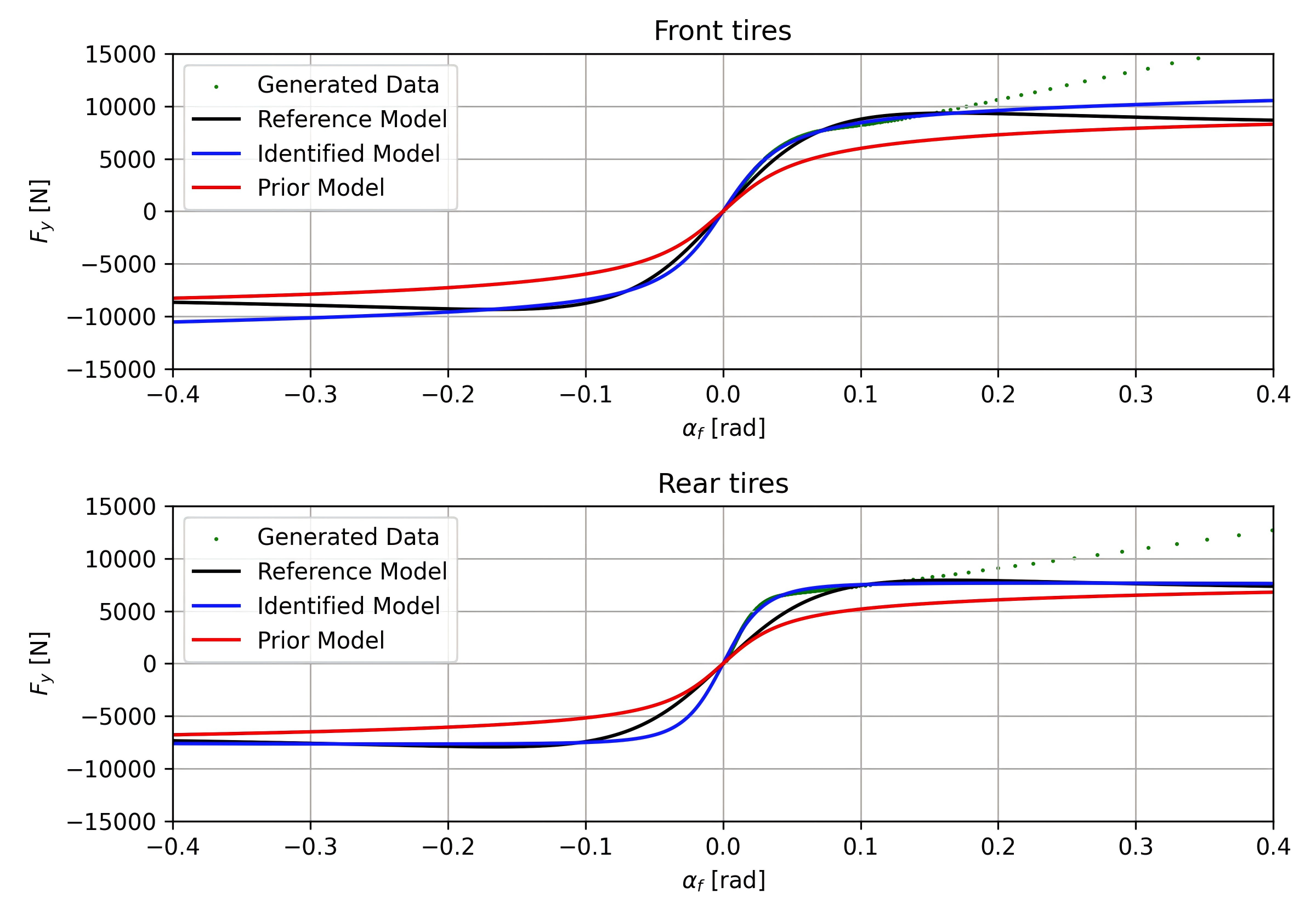}
    \caption{RNN}
    \label{fig:rnn_fit}
  \end{subfigure}
  \hfill
  \begin{subfigure}{0.32\textwidth}
    \centering
    \includegraphics[width=\linewidth]{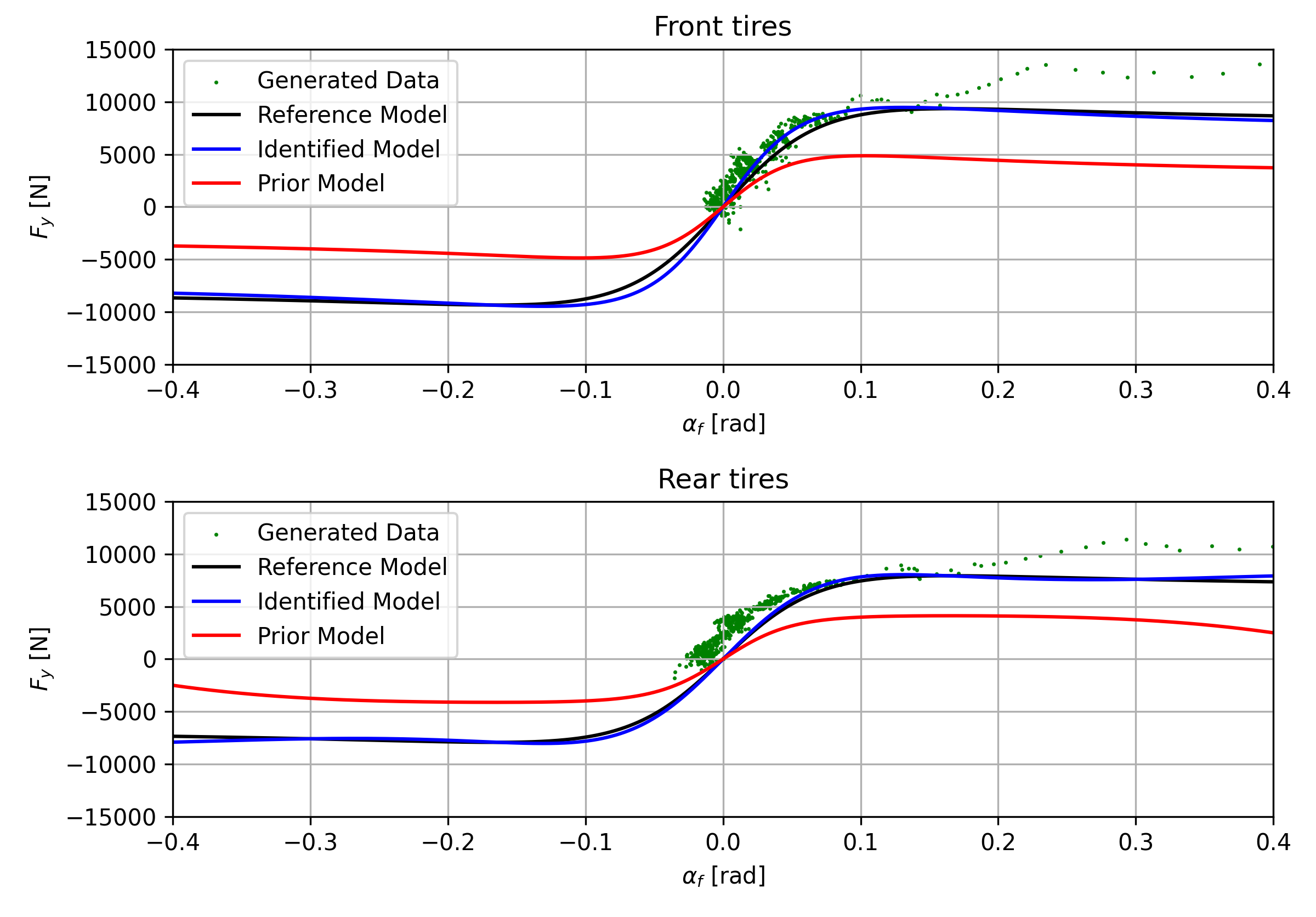}
    \caption{\gls{s4} (Ours)}
    \label{fig:s4_fit}
  \end{subfigure}

  \caption{Comparison of identified Pacejka tire force curves using different neural architectures. The proposed \gls{s4} model achieves the highest identification accuracy and best alignment with the ground truth.}
  \label{fig:architecture_comparison}
\end{figure*}

Beyond scalar metrics, the qualitative physical fidelity of the architectures is visualized in Fig.~\ref{fig:architecture_comparison}, which illustrates the identified lateral force curves ($F_y$ vs. $\alpha$). The \gls{s4} model leverages global convolutions and the \gls{hippo} framework to rigorously capture both high-frequency transients and long-term inertial dependencies. This sequence-level accuracy ensures high-fidelity synthetic data generation, enabling the Nelder-Mead optimizer to precisely and reliably resolve tire saturation limits. Conversely, the MLP and RNN models exhibit visible deviations and underestimations in the critical non-linear handling regions.

\section{Conclusion}
In this paper, a vision-accelerated, sequence-based framework for online system identification in high-speed autonomous racing was presented. By utilizing a MobileNetV3 backbone to generate a probabilistic friction prior, the proposed method effectively mitigates the cold-start initialization issue and significantly reduces parameter convergence time. To capture unmodeled transient dynamics, the \gls{s4} architecture was integrated, demonstrating superior identification accuracy compared to conventional MLP and RNN architectures. Combined with a derivative-free Nelder–Mead optimizer in a virtual steady-state environment, physically interpretable Pacejka tire parameters were reliably extracted.

Extensive co-simulation results demonstrate the framework's effectiveness. The lightweight vision backbone reduces friction estimation error by 76.1\% using 85\% fewer FLOPs, which directly accelerates cold-start convergence by reducing optimization iterations by 71.4\%. Furthermore, by capturing transient dynamics, the \gls{s4} residual model decreases lateral force RMSE by 78.2\% compared to conventional architectures. The resulting framework enables accurate and efficient tire model identification, providing a practical and interpretable solution for real-time autonomous racing applications.

\bibliographystyle{unsrt}
\bibliography{bibliography}

\end{document}